\documentclass[letterpaper, 10 pt, conference]{ieeeconf}  % Comment this line out if you need a4paper

\IEEEoverridecommandlockouts                              % This command is only needed if 
\usepackage{amsmath} % assumes amsmath package installed
\usepackage{amssymb} % assumes amsmath package installed
\usepackage{cite}
\usepackage{subcaption}
\usepackage{amsfonts}
\usepackage{algorithmic}
\usepackage{textcomp}
\usepackage{graphicx,color}
\usepackage{epstopdf}
\usepackage{url}
\usepackage [english]{babel}
\usepackage [autostyle, english = american]{csquotes}
\MakeOuterQuote{"}

\title{A Distributed Hybrid Hardware-In-the-Loop Simulation framework for Infrastructure Enabled Autonomy }

\author{Abhishek Nayak$^{1}$ ~ Kenny Chour$^{1}$ ~ Tyler Marr$^{1}$ ~ Deepika Ravipati$^{1}$~ Sheelabhadra Dey$^{1}$ ~ Alvika Gautam$^{2}$ \\Swaminathan Gopalswamy$^{3}$ ~~ Sivakumar Rathinam$^{3}$ % <-this % stops a space
\thanks{$^{1}$Graduate Student, Texas A \& M University, College Station, TX 77843, U.S.A.}%
\thanks{$^{2}$Graduate Student, IIIT-Delhi, Delhi 110020, India}%
\thanks{$^{3}$Faculty Member, Texas A \& M University, College Station, TX 77843, U.S.A.}%
}

\begin{document}

\maketitle
\thispagestyle{empty}
\pagestyle{plain}

\begin{abstract}
 Infrastructure Enabled Autonomy (IEA) is a new paradigm that employs a distributed intelligence architecture for connected autonomous vehicles by offloading core functionalities to the infrastructure. In this paper, we develop a simulation framework that can be used to study the concept. A key challenge for such a simulation is the rapid increase in the scale of the computations with the size of the infrastructure to be considered. Our simulation framework is designed to be distributed and scales proportionally with the infrastructure.  By integrally using both the hardware controllers and communication devices as part of the simulation framework, we achieve an optimal balance between modeling of the dynamics and sensors, and reusing real hardware for simulation of proprietary or complex communication methods. Multiple cameras on the infrastructure are simulated. The simulation of the camera image processing is done in distributed hardware and the resultant position information is transmitted wirelessly to the computer simulating the autonomous vehicle. We demonstrate closed loop control of a single vehicle following given waypoints using information from multiple cameras located on Road-Side-Units. 
\end{abstract}


% \fi
%
% For peer review papers, this IEEEtran command inserts a page break and
\IEEEpeerreviewmaketitle

\section{Introduction}
Large societal benefits in terms of traffic safety, number of lives saved, number of crashes avoided, fuel savings, and much more have been promised with the advent of autonomous vehicles \cite{EnoPaper2013}. However, such benefits can be reaped only when there is significant penetration of autonomous vehicles in our driving patterns.

Driving functionality can be decomposed into three parts: (i) Situation Awareness (SA) synthesis using one or more sensors to develop a contextual and self-awareness for the vehicle and driver, (ii) Driving Decision Making (DDM) which merges the self awareness information with mission information to define desired motions, and (iii) Drive-by-Wire (DBW) that actually translates the desired motion into actual motion of the vehicle. Conventional vehicles usually take responsibility for the DBW functionality, leaving the human driver to synthesize the SA information and perform DDM. On the other hand, the current generation of autonomous vehicles assumes full responsibility for all three driving functions. This in turns requires the automakers to shoulder the associated higher liability, which in turn can be a big dis-incentive to large scale introduction of autonomous vehicles. Correspondingly, the society might not realize the promise of autonomous vehicles for a very long time.

Infrastructure Enabled Autonomy (IEA) is a new paradigm that addresses the above challenge \cite{IEAConceptPaper} by offloading core functionalities to the infrastructure. Specifically, the synthesis of SA information is offloaded to the infrastructure by mounting smart multi-sensor packs (MSSP's) on Road-Side-Units (RSU's). The MSSP's will monitor the traffic and generate and communicate the SA information with the vehicle by leveraging wireless connectivity. Specialized Smart-Connect (SC) devices embedded in the car receive this communication from the MSSP's, perform DDM, and communicate the decisions locally to the DBW system of the car. 

With the IEA, the OEM takes primary responsibility only for the core DBW system, which they can already produce at large volumes and at very high reliability.  The Infrastructure providers and the SC device makers take the responsibilities for SA synthesis and DDM respectively. Correspondingly, the liabilities get distributed, providing a favorable environment for higher penetration of autonomous vehicles, and society benefiting from the promised safety and economic benefits of AV's.

\begin{figure}[htb]
\centering
\includegraphics[scale=0.8]{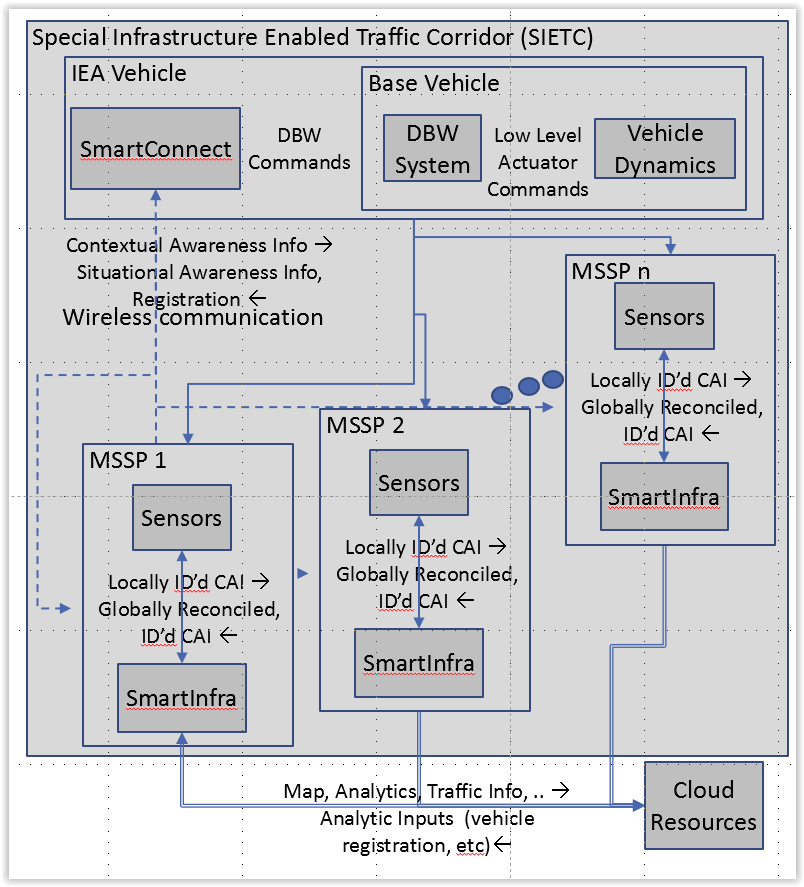}
\caption{IEA Functional Schematic}
\label{fig:IEAFuncSchematic}
\end{figure}

Figure \ref{fig:IEAFuncSchematic} shows a functional view of the IEA concept. There are indeed many open research questions and challenges related to the development of the IEA concept. Some of the immediate challenges in developing the IEA technology would be the complexity, cost, scale, and development safety, which collectively become a huge barrier. Therefore, it is necessary to develop a simulation environment that could be used to develop and evaluate the IEA technologies. 

Developing such a simulation environment is challenging due to its complexity and scale. As the number of RSU's increase, the computational need of the simulation increases significantly. In addition, the communication protocols are already computationally intensive to simulate, and are highly dependent on proprietary hardware characteristics of the equipment manufacturer. This paper describes how we address the above challenge by developing and implementing a hybrid ($i.e.$ mixed models, software and hardware) and distributed (over multiple computers) simulation architecture. 

The scope of the simulation is limited to one vehicle driving down a road equipped with RSUs that each had cameras, computers, and DSRC\footnote{Dedicated Short Range Communication} devices that could communicate to the on-board SmartConnect computer wirelessly. The SmartConnect then interacts with the DBW of the vehicle to drive the vehicle down the road on prescribed paths.The objective of this paper is not develop the control system itself, but rather the ability to support the development of the closed loop control system through a simulation environment. Hardware In the Loop (HIL) simulations have been developed for various driver assistance and vehicle systems in the literature\cite{HIL4,HIL5,HIL1,HIL2,HIL3}; however, the proposed IEA concept is new and this article develops a distributed, decentralized architecture to evaluate the benefits of the proposed concepts.

In section II, we discuss the overall simulation architecture. The sensors used and the vehicle dynamic models are discussed in section III. The vision processing and vehicle control algorithms are presented in sections IV and V respectively. Sections V1 and VII explain the communication and driving vehicle setup in the HIL simulation. The simulation results are presented in section VIII. The article concludes with a summary and future work in section IX.

\section{Distributed Hybrid Simulation Architecture}
\subsection{Simulation Needs}
The simulation needs that are to be addressed can be described in the context of the IEA Functional decomposition, as shown in Figure \ref{fig:IEASimNeeds} below. 
\begin{figure}[htb]
\centering
\includegraphics[scale=0.5]{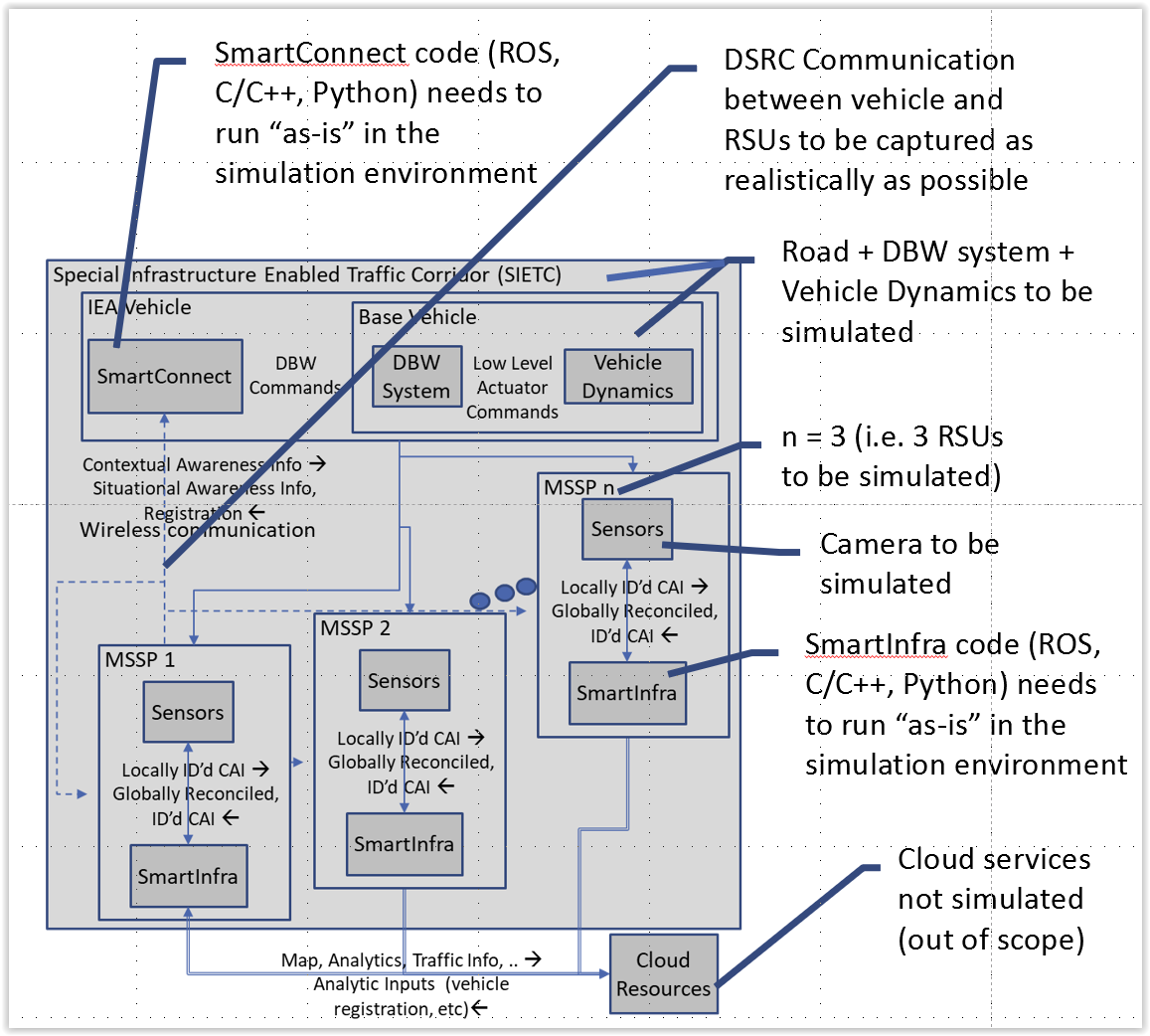}
\caption{Requirements for the IEA Simulation Environment }
\label{fig:IEASimNeeds}
\end{figure}
\begin{enumerate}

\item A key goal of the simulation environment is to enable verification of functionality of control software prior to evaluating in the real vehicle in order to provide safety, decrease the time required to test software changes, and provide greater convenience for testing. This in turn requires that we test the software in the vehicle (SmartConnect) and in the infrastructure (SmartInfra) "as-is" in the simulation environment. 

\item In order to trust the results of the simulation, we need the simulated performance to match the real system as closely as possible. The physical systems that need to be simulated include the vehicle dynamics and the environment (the roads), the sensors (cameras) on the RSUs, and the DSRC communication.

\item We need to simulate the cameras corresponding to each of the MSSPs and deliver a video stream to the respective MSSP that can be processed live. 

\item Since the speed of vision processing is a critical factor in the quality of the overall system performance, it is important that the system be able to run in realistic time frames (close to real-time performance).

\end{enumerate}

\subsection{Simulation Architecture}
To satisfy the first requirement above, we decided to perform the simulation of the SmartConnect (vehicle control) and the SmartInfra in an environment that we expected the real experiments would run in  - namely, in separate computers running Linux OS and ROS. 

In accordance with the second requirement, we decided to simulate the environment using Gazebo and standard vehicle dynamic models. This model runs in the same computer as the vehicle controller to minimize latencies. We will call this computer the Vehicle Computer.

However, the simulation of the cameras on the same computer as the vehicle dynamics poses a huge problem for two reasons: (i) Simulation of multiple cameras very quickly slows down the cameras the system becomes unscalable, and (ii) The camera images need to be streamed to the different MSSPs and this introduces significant bandwidth issues slowing down the simulation and rendering the simulation unscalable. 

Our solution for the above problem was to simulate the cameras on the same computer as the MSSPs (the MSSP Computer). This immediately takes care of the data-streaming problem. The camera will simulate using the same environment and vehicle as the vehicle computer. However, instead of re-simulating the vehicle dynamics, the MSSP computer receives the vehicle information directly from the vehicle computer through wireless means. This also allows the different MSSPs to be synchronized with the vehicle computer.  Since each MSSP computer simulates only its own sensors, the simulation environment is fully scalable. This also addresses requirement 3 above.

Finally, we need to simulate the DSRC communications. In light of requirement 4 and the complexity of DSRC wireless communications, we decided to keep the actual DSRC hardware as part of the simulation environment. In fact, we use the DSRC to also transmit the vehicle information from the vehicle computer to the MSSP computer to facilitate the camera simulation.

\begin{figure}[!ht]
\centering
\includegraphics[scale=0.22]{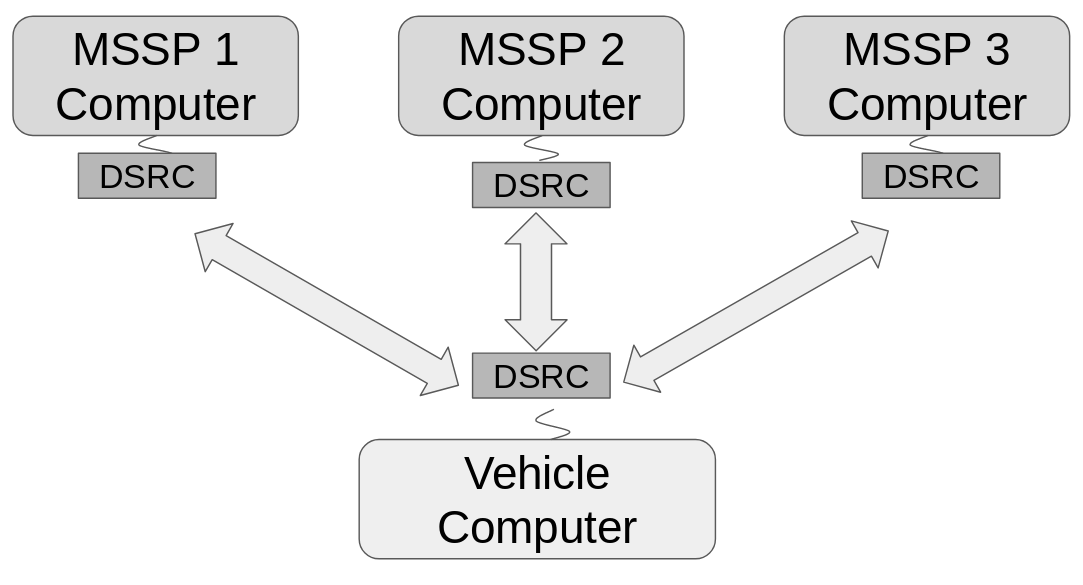}
\caption{Full hardware setup of hybrid simulation}
\label{fig:Simulation_Setup}
\end{figure}

This architecture is captured in Figure \ref{fig:Simulation_Setup}.

\begin{figure}[!ht]
\centering
\includegraphics[scale=0.23]{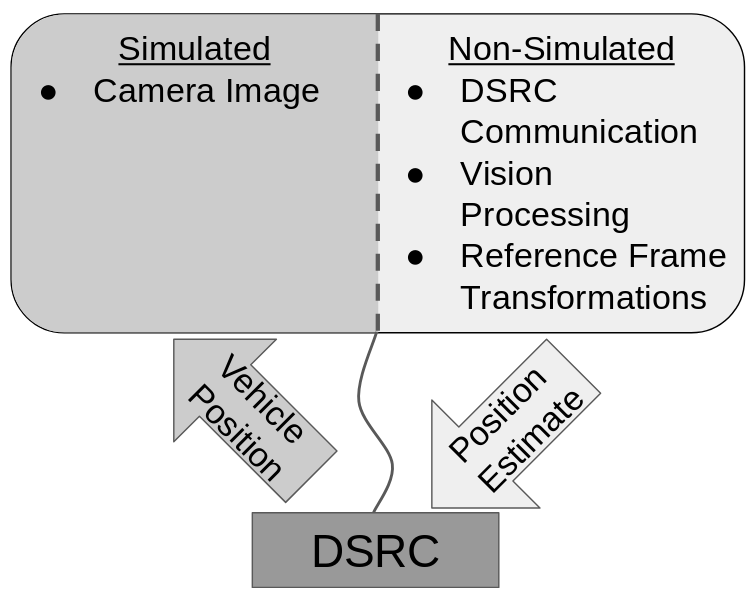}
\caption{Detailed setup of MSSP computer processes}
\label{fig:MSSP_Computer}
\end{figure}
\begin{figure}[!ht]
\centering
\includegraphics[scale=0.23]{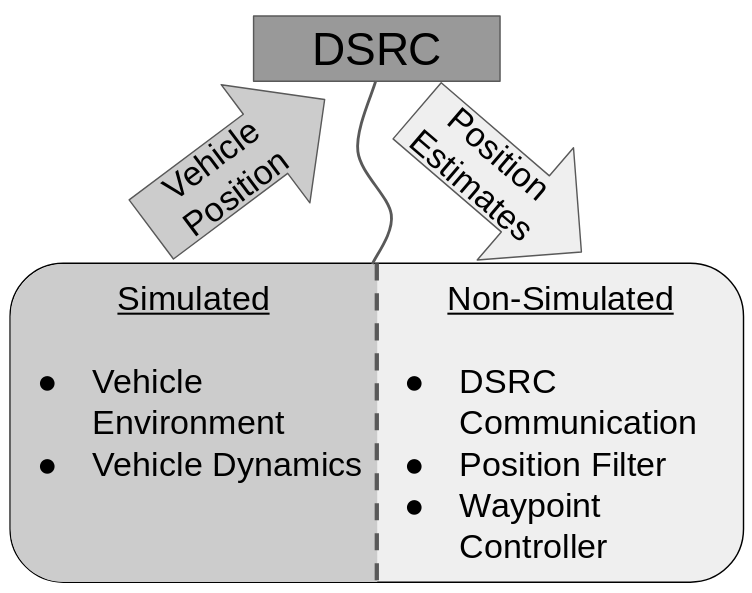}
\caption{Detailed setup of vehicle computer processes}
\label{fig:Vehicle_Computer}
\end{figure}

\subsection{Simulation Setup}
The current hybrid simulation setup is composed of four total computers, representing an equivalent span of road covered by three MSSP units. One computer is dedicated per MSSP system with an accompanied camera, and the last computer acts as the vehicle’s on-board computer. Each computer is connected to a DSRC unit in the same manner as the real system and this is the only means of communication between all of the computers. Upon launching the simulation, each computer launches a Gazebo simulation along with all of the same software that will be used in the full experimentation in accordance with the computer’s role. Pictured in Fig. \ref{fig:MSSP_Computer}, the MSSP computer runs as it would in the full system, with the exception of the camera image being simulated. In order to accurately simulate the camera image the computer receives communication of the simulated vehicle position via the DSRC unit, and replicates this exact same position in its own simulation environment. Once this image is simulated, the rest of the operations occur as they would in the real system and the computer returns a position estimate based off of vision processing. Represented in Fig. \ref{fig:Vehicle_Computer}, the vehicle computer simulates the vehicle environment and dynamics in addition to its usual processes. 
This distributed hybrid simulation allows for testing that closely resembles the real life system. By minimizing the simulated aspects of the testing, this architecture will provide results that should resemble full experimentation.
%Accurate vehicle dynamics are used along with a simulated camera that is positioned the same as the real system. The remainder of this hybrid simulation uses the same hardware that will be used in full experimentation. In addition, all of these computers are running Gazebo simulations which increases the overall computational load. Therefore it can be expected that the computer processing in the real system will be accomplished at a similar or faster pace than that which is currently experienced. These factors together indicate that this architecture will provide results that should resemble full experimentation. \\

\section{Sensors and Vehicle Dynamics Models}
A simulation model of Lincoln MKZ vehicle provided by Dataspeed ADAS \cite{url_dataspeed} is used for the vehicle dynamic model in this paper. The vehicle, the environment sensors, as well as the road environment are all modeled in gazebo. The vehicle platform is equipped with sensor modules such as GPS and an Inertial Measurement Unit (IMU). 

Communication with the vehicle controller happens on ROS through DSRC, where sensor data can be received from, and desired actuation, velocity commands can be sent to the vehicle as ROS topics. 

The vehicle also has a DBW system that communicates with the SmartConnect through ROS. The DBW receives velocity and steering commands and passes that on as appropriate actuation commands (throttle, steering, gear and brake) to  drive the vehicle. The vehicle sensor data such as position can be read by the SmartConnect through ROS.

The vehicle model follows a bicycle model for lateral and longitudinal vehicle kinematics and dynamics \cite{vehicle_dynamics} which takes into account vehicle motion characteristics at both slow and high speeds.
\begin{figure}[!ht]
\centering
\includegraphics[scale=0.40]{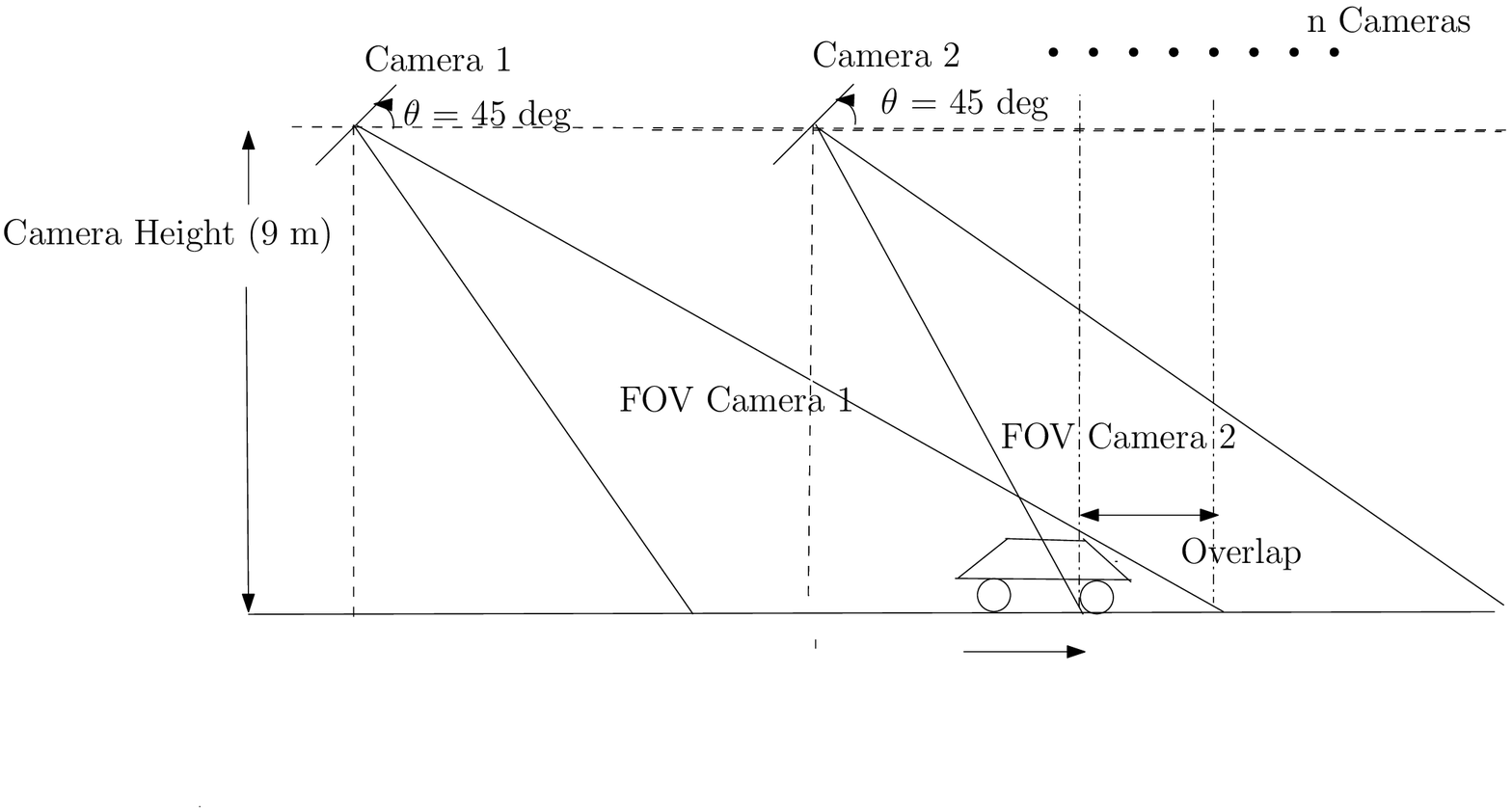}
\caption{Simulation setup demonstrating the successive camera placements with overlap}
\label{fig:sim_setup}
\end{figure}

\begin{figure}[!ht]
\includegraphics[scale=0.30]{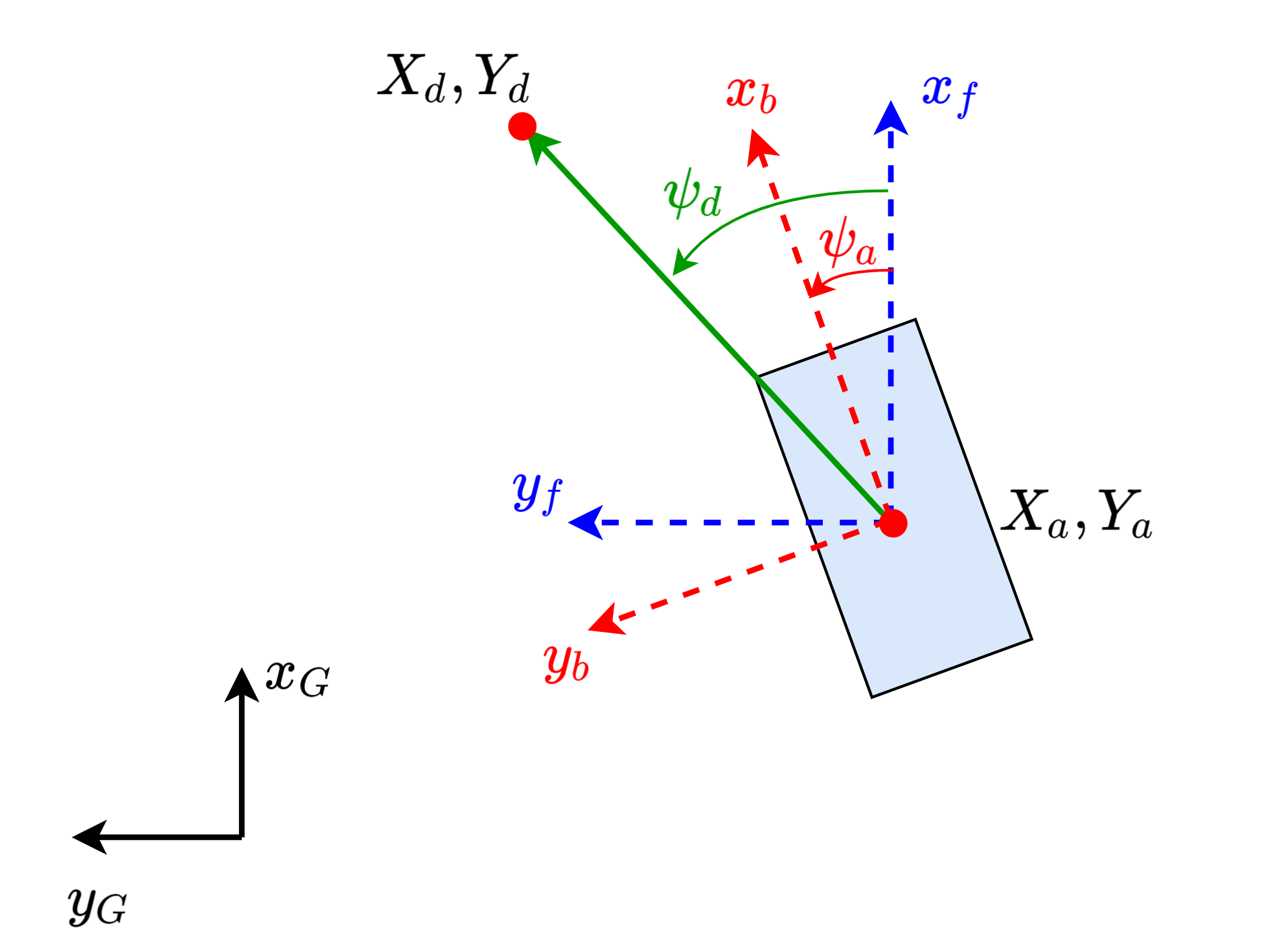}
\caption{Simulation reference frames}
\label{fig:frames_setup}
\end{figure}

A variety of road environments can be setup using the modular road segment gazebo models provided as a part of the simulator. For the initial testing, we consider a straight line road scenario. The cameras on the road are modeled using the sensor plugins available in gazebo, which uses the standard pinhole camera model with the assumption of no distortion. The camera parameters like intrinsic matrix and distortion parameters are available as a ROS topics. The cameras are initialized at a fixed position and orientation in the gazebo world frame with some overlap between successive cameras. Fig. \ref{fig:sim_setup} shows the camera setup in the gazebo simulator. The cameras are mounted at an altitude of $9$ m and orientation of each camera is ($\phi=0 \deg,\theta=45 \deg, \psi=0 \deg$) with respect to horizontal. The figure also shows the overlapping region between the two cameras which is where the handover of control takes place from one host MSSP to next. This will be discussed in detail in subsequent sections. The vehicle position is evaluated from these road cameras which is then used along with the vehicle heading from the IMU to generate appropriate control commands.\\
Reference frames and conventions, used through out the paper are shown in Fig. \ref{fig:frames_setup} ,
.\\
\begin{itemize}
\item $x_G$, $y_G$ := Gazebo World Frame
\item $x_b$, $y_b$ := Body-fixed frame
\item $x_f$, $y_f$ := Intermediate body-fixed, non-rotating frame
\item $X_a$,$Y_b$ := Vehicle position in gazebo world frame
\item $X_d$,$Y_d$ := Target waypoint in gazebo world frame
\end{itemize}
\section{Vision Processing}
The goal of vision processing is to detect and track the car, and publish the image coordinates of a bounding box that tightly encloses the car in a ROS topic. These image coordinates are later converted to world coordinates in the MSSP. Naturally, it is important to accurately track the vehicle as this directly affects the performance of the overall system.  

Car detection and tracking is accomplished by integrating Background subtraction\cite{BGsubtract} and Tracking-Learning-Detection(TLD) \cite{tldZdenek}  algorithms. A background subtraction algorithm is first used to identify the presence of a car in the images. This algorithm is relatively simple to implement and essentially compares the current image with the previous images to find any significant shifts in the intensity values of the pixels. As the cameras are stationary, any significant discrepancies must imply the entry of new objects or the motion of already existing objects. In the HIL setup, the car was the only moving object and therefore, the background subtraction algorithm could readily find the car. Once the car is identified in an image, its coordinates then provide the cue for the tracking algorithm which is performed using the TLD. We use TLD primarily because it is fast and widely used in tracking applications. In the event when TLD loses track of the car, the background algorithm is started again and the process is repeated. 

TLD \cite{tldZdenek} is an algorithm for long-term tracking of unknown objects in a video without any training process. The algorithm is fast enough to track objects even in high frame-rate videos. The TLD algorithm is a unique composite of tracking, learning and detection components. While the tracker estimates the object position between the frames, the detector scans through each frame and localizes the object. The learning component of the algorithm keeps observing the performance of tracker, detector and estimates the detector's error. It also improves the detector's performance by generating positive examples and training the detector online. The next section discusses the transformation procedure from the image coordinates of the bounding box representing the car to global coordinates.

\subsection{Transformation}
\begin{figure}[!ht]
\centering
\includegraphics[scale=0.40]{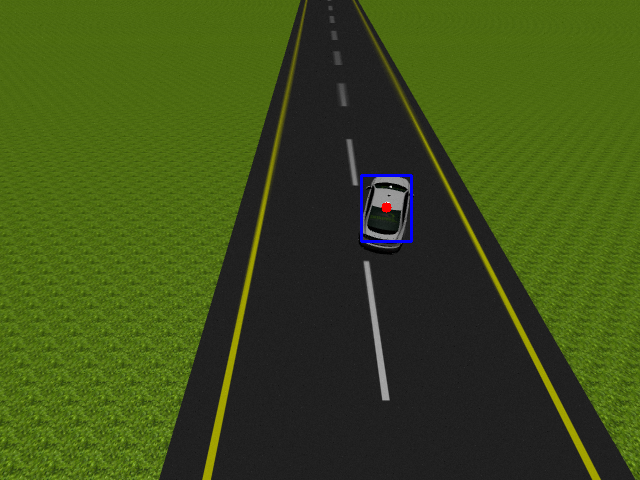}
\caption{A processed image showing the bounding box which encloses the car.}
\label{fig:car_detection}
\end{figure}
The center of the vehicle bounding box as output by the vision processing block is used as an input for transforming the vision estimates to global position coordinates relative to the MSSP frame. Fig. \ref{fig:car_detection} shows the view from one of the cameras along with the car detected in the frame.

Let $\left[x,y,1\right]$ be the homogeneous image  coordinates of a pixel and $\left[X,Y,Z,1\right]$ be corresponding vehicle position on the road relative to the MSSP in homogeneous coordinates. The mapping between the two is given by,
\begin{eqnarray}\label{eq:im_world_mapping}
\left[ \begin{array}{c} x \\ y \\ 1 \end{array} \right] = K \begin{bmatrix} R ~t\end{bmatrix} \times \left[ \begin{array}{c} X \\ Y \\ Z \\ 1 \end{array} \right]
\end{eqnarray}
where $K$ is the intrinsic camera matrix, $R$ is the derived rotation matrix from camera's orientation and $t$ is the translation vector derived from camera's known orientation and position. Let $\left[M ~ p_4\right]$ = $K \times \left[ R ~t\right]$ be the 3x4 camera matrix with the 3x3 sub-matrix $M$ and column vector $p_4$. The back-projection equation to transform the image pixel coordinate to world points \cite{transform_ref} is given by 
\begin{eqnarray}
P = C + \lambda M^{-1} p
\end{eqnarray}
Here, $p$ is the 2D homogeneous column vector $[x ~y ~1]'$, $C$ is the camera center $(0,0)$ in non-homogeneous world coordinates (given in homogeneous coordinates by $C = -M^{-1} p_4)$, 
and lambda is an arbitrary scalar that controls the depth. This will give the non-homogeneous 3D point $P$. To back-project to an exact depth $d$, we set $\lambda = d||m_3||$ where $m_3$ is the third row of $M$, and  $d$ is the camera altitude. This gives the position of the vehicle relative to the MSSP.

\section{Vehicle Control}
%Since the purpose of this work is demonstration of the simulation environment for IEA, we used a simple way-point controller.

\subsection{Desired Trajectory Generation}

In a typical IEA operation, the SmartConnect will receive global planning routes, from which it will perform local path planning. For this demonstration of the simulation environment, the local path planning was just a linear interpolation of the (coarse) global path to generate a finer local path. This path planning communication happens at a much lower frequency than the closed loop control. Since we have only 3 MSSPs in this simulation, the generation of the local path is done at the beginning, prior to closed loop control.

The way-points are collected into an array as a series of coordinates \textbf{($Xd_{i}, Yd_{i}$)} that represent the desired vehicle path in the global reference frame. A look-ahead distance \textbf{$d$} from the vehicle’s current position and the next waypoint in sequence determines the next coordinate for the vehicle to approach \textbf{($Xd, Yd$)} at each loop.
\begin{figure}[!ht]
\centering
\includegraphics[scale=0.50]{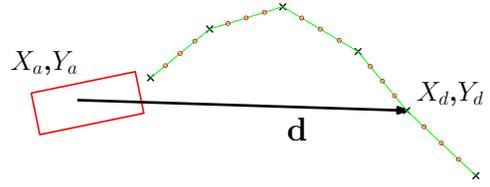}
\caption{Vehicle lookahead distance with interpolated path}
\label{fig:look-ahead}
\end{figure}\\ 

\subsection{Controller}
\begin{figure}[!ht]
\centering
\includegraphics[scale=0.17]{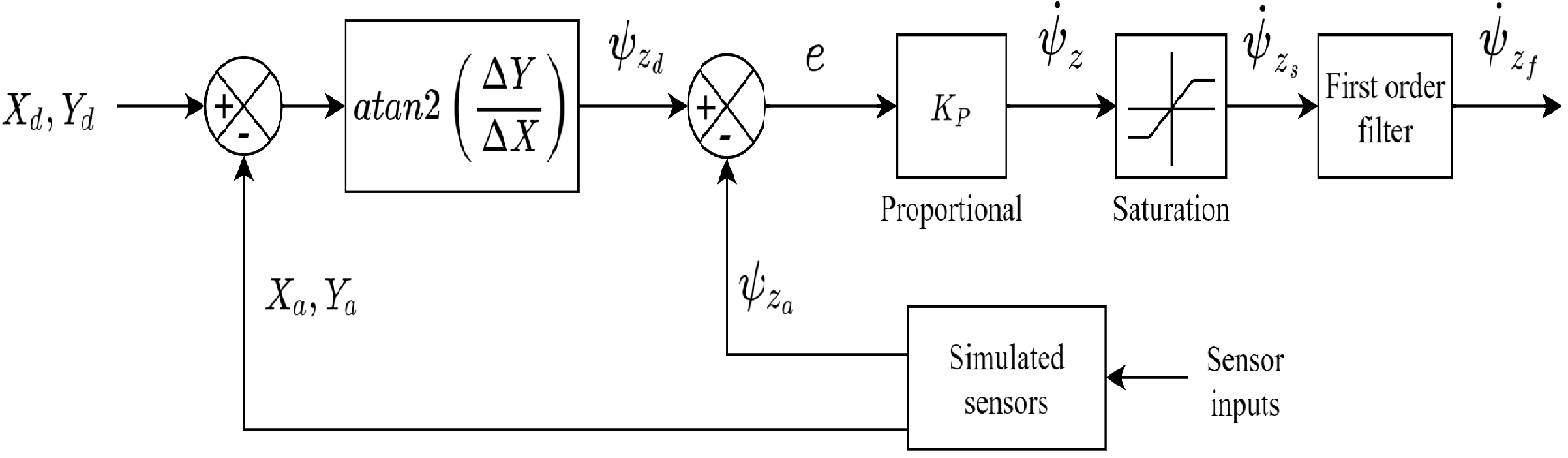}
\caption{Vehicle controller block diagram}
\label{fig:Vehicle_controller}
\end{figure}

The vehicle's yaw angle is controlled to minimize heading error, while a constant linear forward velocity is provided.  Inputs for the vehicle controller are the following: (i) next target waypoint coordinate ($X_{d}, Y_{d}$) (ii) current vehicle yaw-angle measurement $\psi_{z_{a}}$ and (iii) current vehicle position measurement ($X_{a}Y_{a}$). The resulting output is a filtered yaw-rate command $\dot{\psi_{z_{f}}}$ which steers the vehicle. The controller consists of two loops to provide two types of errors. The outer loop is used to generate the desired heading angle, based on the difference between the vehicle’s desired and current position. The inner loop is used to compute the vehicle heading error by calculating the difference between the desired heading from the previous loop and the current yaw angle measurement. %Since both of these angle measurements are with respect to the $X_{f}$ axis, this is possible.
The yaw rate command is the product of a proportional gain $K_p$ and heading error $e$ which is further processed in the remaining blocks. A controller structure is shown in figure \ref{fig:Vehicle_controller}. 
\\ 
The initial output yaw-rate from the proportional controller enters a saturation function block to prevent excessive command velocities. Next, a first-order non-recursive filter is used to smooth the angular velocity commands to the vehicle. 

\begin{eqnarray}
\dot{\psi}_{z_{f}}^{k+1}=\alpha\dot{\psi}_{z_{f}}^{k}+(1-\alpha)\dot{\psi}_{z_{f}}^{k-1}
\end{eqnarray}

More weighting is given to the previous vehicle yaw-rate command velocity rather than the current output to avoid sudden changes. This is done by selecting a small $\alpha$. The tunable parameters in this system include (i) proportional heading error gain (ii) output saturation limits and (iii) output filter weight. 

\section{Communication}
\begin{figure}[!ht]
\centering
\includegraphics[width=3.5in]{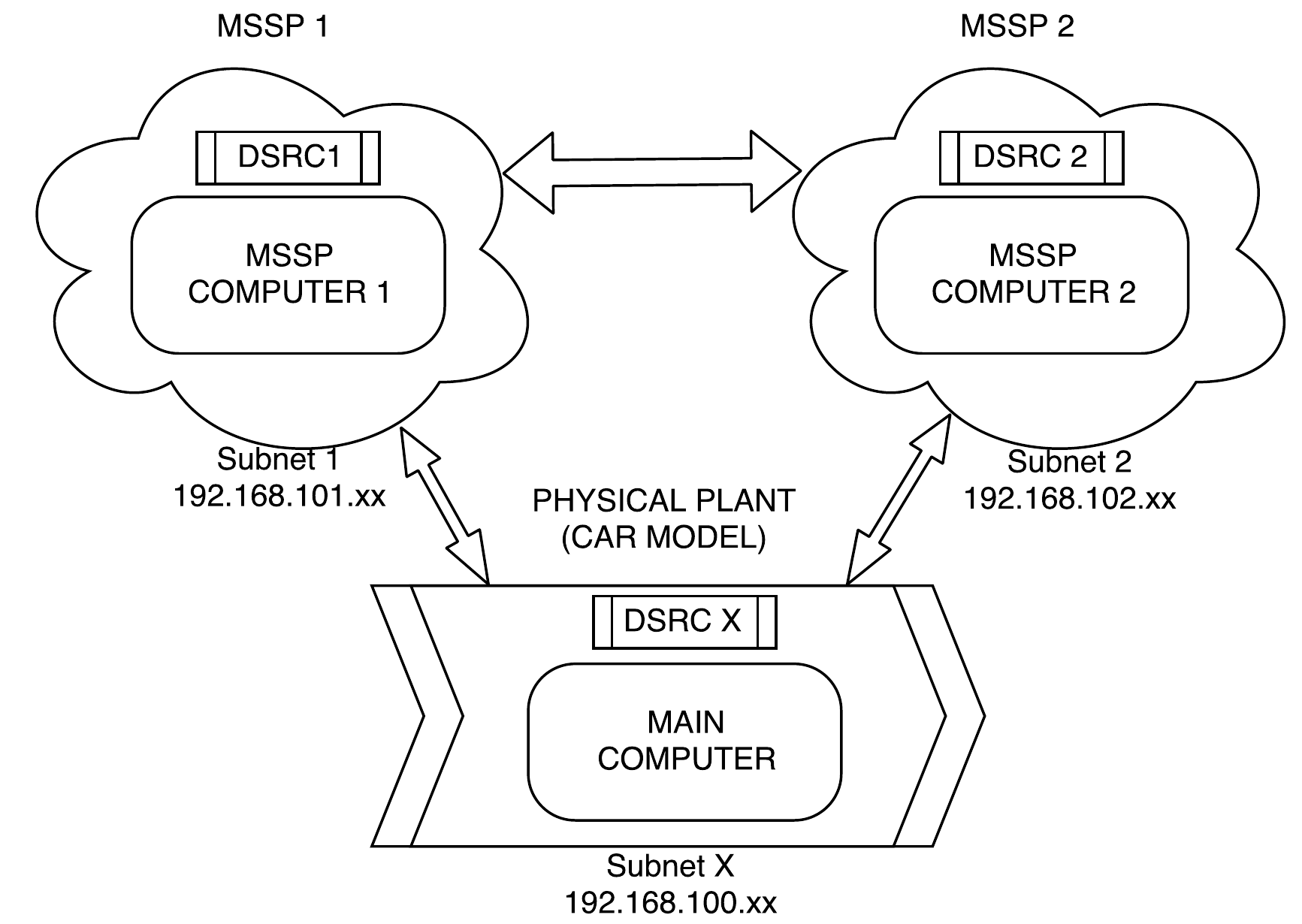}
\caption{Network layout of Car \& MSSP units}
\label{fig:IEA_layout}
\end{figure}
Actual DSRC units are used to establish communication between different MSSP computers and the main computer containing the physical plant model simulation as part of the simulation setup forming a Hardware-in-loop setup. The DSRC units used were WSU 5001 supplied by DENSO International America Inc. Each MSSP unit forms a private network separately and is assigned a unique IP address in IPv4 configuration\cite{rekhter1996address}. This unique IP address is used as an identifier for establishing communication and in turn as a parameter of the defined global zone for each of the MSSP’s. The current network configuration was set-up such that each of the MSSP's (containing the DSRC's and a MSSP computer) were assigned unique IP addresses in separate subnets in 192.168.0.0/16 IP range in-order to make the IP address identification simple ($e.g.$ all the devices in MSSP 2 were assigned IP addresses in 192.168.102.x range). The DSRC units were used as the medium to transmit relevant physical system parameters and variables which form the feedback input to the vehicle controller. Figure \ref{fig:IEA_layout} shows the Network layout when the Physical plant model is communicating with 2 MSSP's. %As we will discuss further in Driving cell management section of this paper, DSRC units formed an integral part of the simulation such that relevant ROS topics were identified and subscribed to on the basis of which DSRC the main car computer was communicating, establishing a closed loop control.

\section{Driving Cell Management}
Communication in the system is accomplished through the ROS \textit{multimaster\_fkie} package. This provides a platform for creating a local area network of several ROS Masters, which allow for the seamless synchronization of nodes between different machines. Separate machines are exposed to each other via hardwired DSRC connections and also explicitly setting the ROS\_MASTER\_URI, ROS\_HOSTNAME, ROS\_IP environmental variables of each machine to the DSRC gateway addresses. All topics between synchronized nodes being published by the MSSPs are known by the vehicle computer prior to the start of the simulation. Once the simulation is started, the vehicle computer listens for all of the position topics published by the MSSP computers. The MSSP computer only publishes a topic to the vehicle if the corresponding camera for that particular MSSP is currently tracking the vehicle. Thus, the vehicle computer does not require its location within the simulation; it simply listens for a topic to be published from one of the MSSP computers. Once given its initial position, the vehicle computer then sends commands to the vehicle based off of its position.

As the vehicle travels, there is a seamless transition between effective areas of the MSSPs, these areas are known as cells. A filter is implemented on the vehicle that allows it to listen to any number of MSSPs simultaneously. This is imperative to the process because cells overlap at their beginning and ends. While the vehicle is in only one cell the position from that particular MSSP is relayed to the controller, however if the vehicle is reported to be in multiple cells, a filter is applied that takes the average of both reported positions.

\section{Simulation Results}
In this section, we present the simulation results and discuss the performance of the proposed architecture with respect to network load, comparison of vision position estimates to true positions, followed by the performance of closed loop vehicle control.
\subsection{Simulation Setup}
This simulation was set up with three cameras placed along a straight road. The cameras were stationary and were placed $40$ meters apart with an overlap of about 13 meters between each of the cameras.
% All he cameras were placed at a y-coordinate of XXX.XX and the x-coordinates were XXX.XX, XXX.XX, and XXX.XX for MSSPs one, two, and three, respectively. 
Two trials were then performed at a speed of $3$ m/s and $6$ m/s respectively. The vehicle was given an initial velocity with a straight forward heading until the it enters the field of view of the camera and vision processing starts. The simulation was then allowed to run until the vehicle was out of frame from the final MSSP, at which point the vehicle automatically stopped. 
\begin{figure}[!ht]
\centering
\includegraphics[scale=0.3]{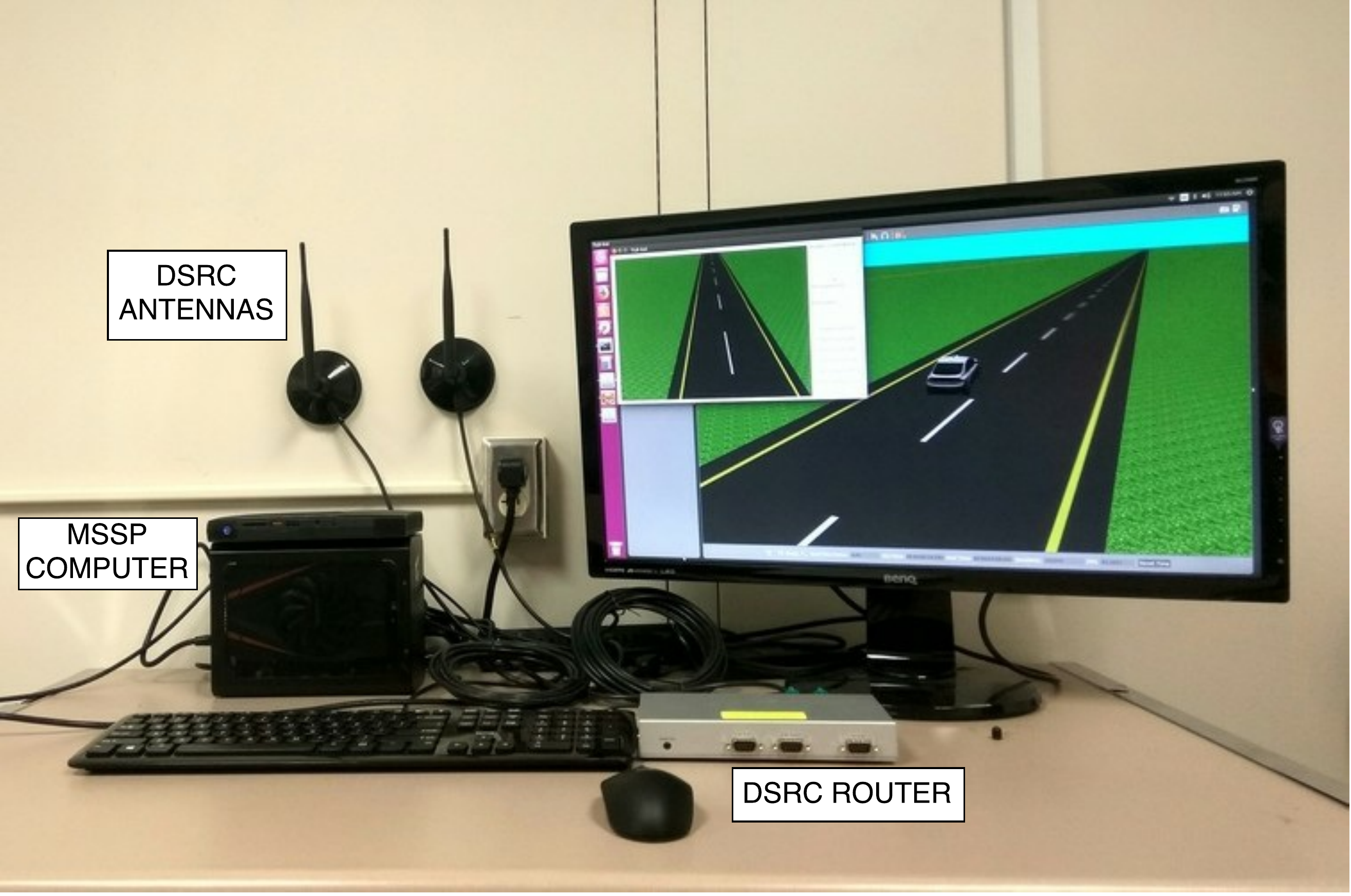}
\caption{Simulation Setup with DSRC Units}
\label{fig:IEA_Setup}
\end{figure}
\subsection{Network Load analysis}
\begin{figure}[!ht]
\centering
\includegraphics[scale=0.4]{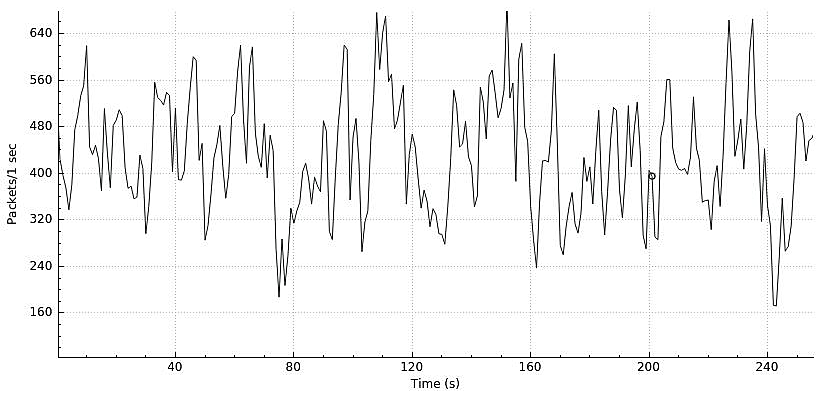}
\caption{Plot of Packet Transmission Rate vs Time}
\label{fig:Packet_rate}
\end{figure}

\begin{figure}[!ht]
\centering
\includegraphics[scale=0.4]{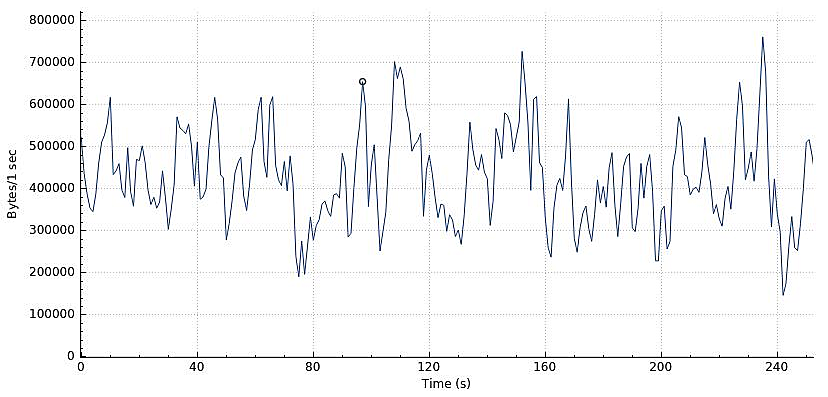}
\caption{Plot of Data Transfer Rate vs Time}
\label{fig:Data_transfer}
\end{figure}

In applications of communication where active safety is involved, investigations into network parameters like network load, data transfer rate, range etc. are of paramount importance. In the current setup, latency was observed to be in the range of 1.5 to 2.0 ms throughout the network operational range with a TxRate specification of 32Mbps for the DSRC routers. Figure \ref{fig:Packet_rate} shows a plot of data packet transmission rate against time. Figure \ref{fig:Data_transfer} shows a plot of data transfer rate against time. The data in these plots correspond to the communication load observed on a single MSSP computer which is synchronized with 2 other MSSP computers and the main computer containing the physical plant model simulation. As the simulation experiment was conducted in a closed room environment, all the MSSP's (DSRC routers) were in-range and synchronized throughout the experiment. In real-outdoor scenarios, the vehicle computer would normally be synchronized with only one of the MSSP's ($i.e.$ its current home MSSP) at any given point of time, and hence, the network load is expected to be lower than what was observed during simulation. 

\subsection{Vision based Situation Awareness Synthesis}
\begin{figure}[!ht]
\centering
\includegraphics[width=3.5in]{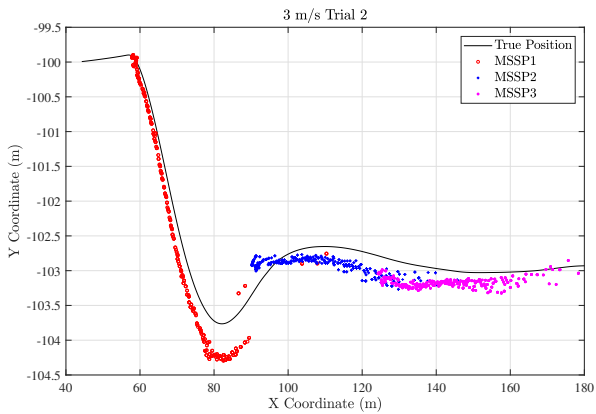}
\caption{True Position vs. MSSP Position Estimates at 3 m/s}
\label{fig:Odom_Mssp_3T2}
\end{figure}

\begin{figure}[!ht]
\centering
\includegraphics[width=3.5in]{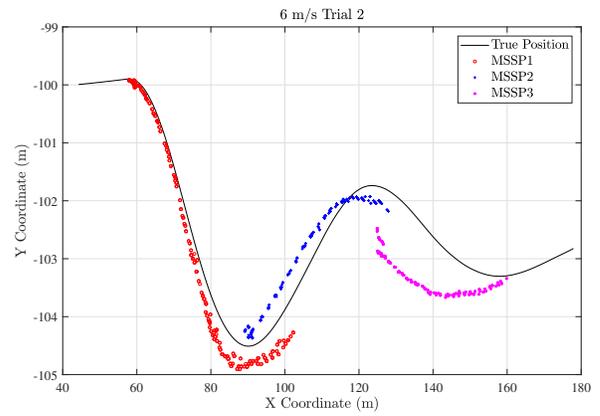}
\caption{True Position vs. MSSP Position Estimates at 6 m/s}
\label{fig:Odom_Mssp_6T2}
\end{figure}

Throughout the simulations, the true positions of the vehicle and their estimates from each MSSP are recorded. This data can be seen in Figure \ref{fig:Odom_Mssp_3T2} and Figure \ref{fig:Odom_Mssp_6T2}. From these figures, it can be seen that the MSSPs are able to closely estimate the true position of the vehicle at all times.


\subsection{Cell Management}
\begin{figure}[!ht]
\centering
\includegraphics[width=3.5in]{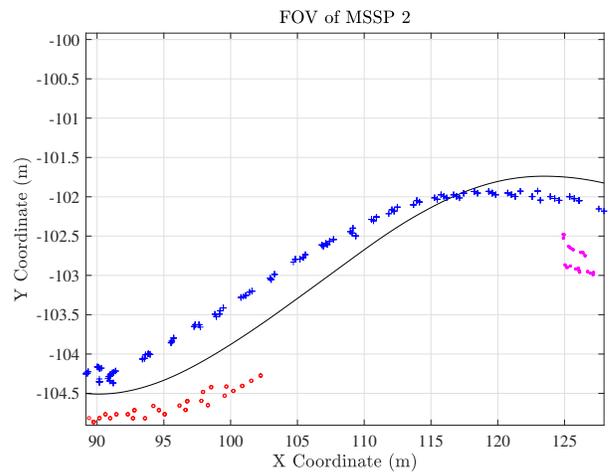}
\caption{True Position vs Estimate at 6m/s for Camera 2}
\label{fig:Odom_MSSP_V6T2_Cam2}
\end{figure}

The cell management can also be seen in figures \ref{fig:Odom_Mssp_3T2}, \ref{fig:Odom_Mssp_6T2}.
%images/Odom_vs_MSSP_V3T2.fig and images/Odom_vs_MSSP_V6T2.fig 
From these graphs, the overlapping nature of the different MSSP zones can be observed. A more focused section of data from the 6m/s trial can be seen in Fig. \ref{fig:Odom_MSSP_V6T2_Cam2}
%images/Odom_vs_MSSP_V6T2_Cam2.fig
This graph focuses on only the section of road the MSSP 2 sees. The overlapping of zones 1 and 3 with zone 2 can be seen more clearly from this graph. It can also be seen that in each overlapping sections, there is a noticeable difference between the $y$-coordinate estimations of the two cameras. 
\subsection{Closed Loop Vehicle Control}
\begin{figure}[!ht]
\centering
\includegraphics[width=3.7in]{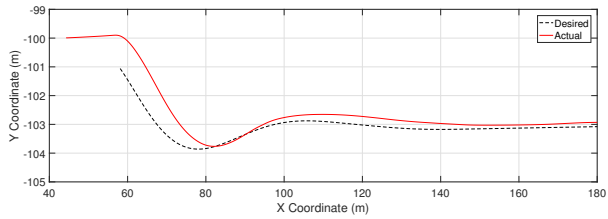}
\caption{Closed-Loop Performance at 3m/s}
\label{fig:closed_loop_3V2}
\end{figure}

\begin{figure}[!ht]
\centering
\includegraphics[width=3.7in]{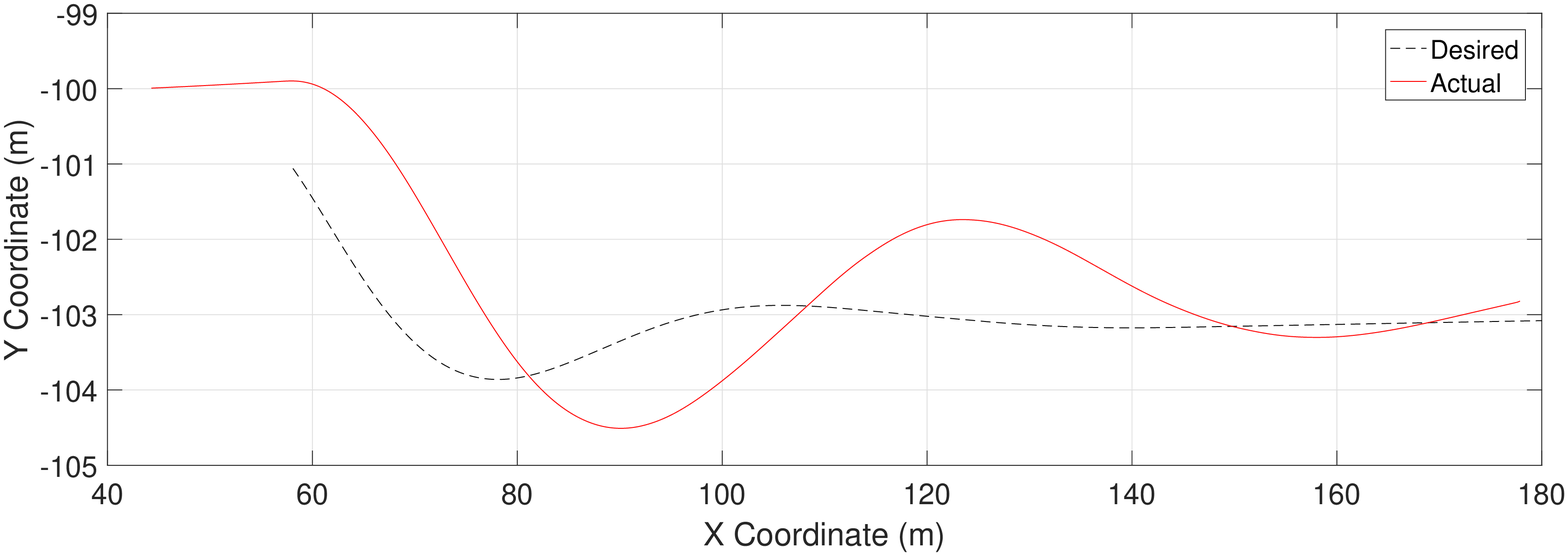}
\caption{Closed-Loop Performance at 6m/s}
\label{fig:closed_loop_6V2}
\end{figure}

The vehicle performed waypoint navigation with sensor feedback from several MSSP's. Results can be seen in Figures \ref{fig:closed_loop_3V2}, \ref{fig:closed_loop_6V2} at different forward linear velocities (3m/s and 6m/s respectively) to ascertain close-loop performance. A baseline for comparison was provided by having the vehicle perform waypoint navigation with perfect sensor information from a single Gazebo simulation alone (representing perfect sensors). 

In both figures, the vehicle travels forward to reach the first MSSP cell in order to obtain feedback. This occurs at approximately 60m on the x-coordinate, where the vehicle begins to head in the direction of the first waypoint that is at least 10m ahead. From figure  \ref{fig:closed_loop_3V2}, the vehicle tracks the desired path reasonably well at steady state. It reaches its target path with little overshoot and settles near the steady state value of the desired trajectory. In figure \ref{fig:closed_loop_6V2}, the vehicle however experiences greater oscillations as it overshot the desired path. These oscillations are expected for a system with sub-optimal parameters or delays as is the case with this project. Relaxation of the saturation limits and more weighting on current output velocity may improve system performance by increasing bandwidth. Additionally, derivative action or lead compensation may also be incorporated to improve system response in terms of rapid changes.

\section{Conclusion}
We have developed and successfully demonstrated a distributed hybrid simulation framework that can be used for the development of IEA. As the controllers on the vehicle (SmartConnect) and the infrastructure (SmartInfra) are both based on C/C++/Python running on ROS and Linux, we leverage the hardware that runs these controllers for the simulation also, thereby enabling "as-is" simulation of the actual control software. The vehicle dynamics and the environment are simulated on the SmartConnect computer, while the cameras are simulated on the SmartInfra computer. This distribution naturally allows for scalability, as additional vehicles or RSUs will automatically be handled by using the respective computers on the vehicles or the RSUs. 

We use the actual DSRC devices in-the-loop to perform the wireless communications. The DSRCs are used for (i) facilitating the communication between the SmartInfra and the SmartConnect, and (ii) synchronizing the simulation of the cameras across the MSSPs. Since we are using real hardware, the SmartInfra-SmartConnect communication is as close to reality in the simulation. For the synchronization, we have identified the typical delays to be of the order of less than 2 ms in the simulation environment inside the lab, and we believe that this delay is negligible in comparison to the scalability provided by the distributed architecture. %However, if this ever becomes a major issue in the future, the synchronization can be attempted through other faster communication means, such as hard-wired cable connections.

We were able to demonstrate a successful closed-loop simulation of a vehicle moving down a straight road but making a lane change, controlled only based on the feedback from cameras on the road. We believe that this demonstrates the feasibility of the overall IEA concept, providing us with the confidence to continue to demonstrate the feasibility experiments in real hardware.


\section*{Acknowledgment}

The authors would like to thank very useful discussions on the conceptual design of the simulations and experiments with Austin Birch, Mike Ashley, Sai Vemprala, Drs. Srikanth Saripalli and Swaroop Darbha, from the Department of Mechanical Engineering, CAST Program, Texas A\&M University.



%\bibliographystyle{IEEEtran}

\end{document}